\documentclass{article}
\usepackage{amssymb}

\usepackage[final]{corl_2025} 
\usepackage{amsmath}
\usepackage{enumitem}
\usepackage{booktabs}
\usepackage{graphicx} 
\usepackage{wrapfig} 
\usepackage{algpseudocode}
\usepackage{algorithm}
\usepackage{multicol}
\usepackage{multirow}

\title{MimicFunc: Imitating Tool Manipulation from a Single Human Video via Functional Correspondence}

\author{
  Chao Tang\textsuperscript{1,2} 
  \quad 
  Anxing Xiao \textsuperscript{2}
  \quad 
  Yuhong Deng \textsuperscript{2}
  \quad 
  Tianrun Hu \textsuperscript{2}
  \quad 
  Wenlong Dong \textsuperscript{1}
  \\[5pt] 
  \textbf{Hanbo Zhang} \textsuperscript{2}
  \qquad
  \textbf{David Hsu} \textsuperscript{2}
  \qquad
  \textbf{Hong Zhang} \textsuperscript{1}
  \\[5pt]
  Southern University of Science and Technology\textsuperscript{1} $\quad$  National University of Singapore\textsuperscript{2} 
  \vspace{-5mm}
}

\begin{document}
\maketitle


\begin{abstract}
    Imitating tool manipulation from human videos offers an intuitive approach to teaching robots, while also providing a promising and scalable alternative to labor-intensive teleoperation data collection for visuomotor policy learning. While humans can mimic tool manipulation behavior by observing others perform a task just once and effortlessly transfer the skill to diverse tools for functionally equivalent tasks, current robots struggle to achieve this level of generalization. A key challenge lies in establishing function-level correspondences, considering the significant geometric variations among functionally similar tools, referred to as intra-function variations. To address this challenge, we propose MimicFunc, a framework that establishes \textit{functional correspondences} with function frame, a function-centric local coordinate frame constructed with keypoint-based abstraction, for imitating tool manipulation skills. Experiments demonstrate that MimicFunc effectively enables the robot to generalize the skill from a single RGB-D human video to manipulating novel tools for functionally equivalent tasks. Furthermore, leveraging MimicFunc's one-shot generalization capability, the generated rollouts can be used to train visuomotor policies without requiring labor-intensive teleoperation data collection for novel objects. Our code and video are available at \href{https://sites.google.com/view/mimicfunc}{https://sites.google.com/view/mimicfunc}.
\end{abstract}

\keywords{Tool Manipulation, Imitation from Human Video} 


\begin{figure}[h]
\centering
  \vspace*{-0.4in}
\includegraphics[width=1\linewidth]{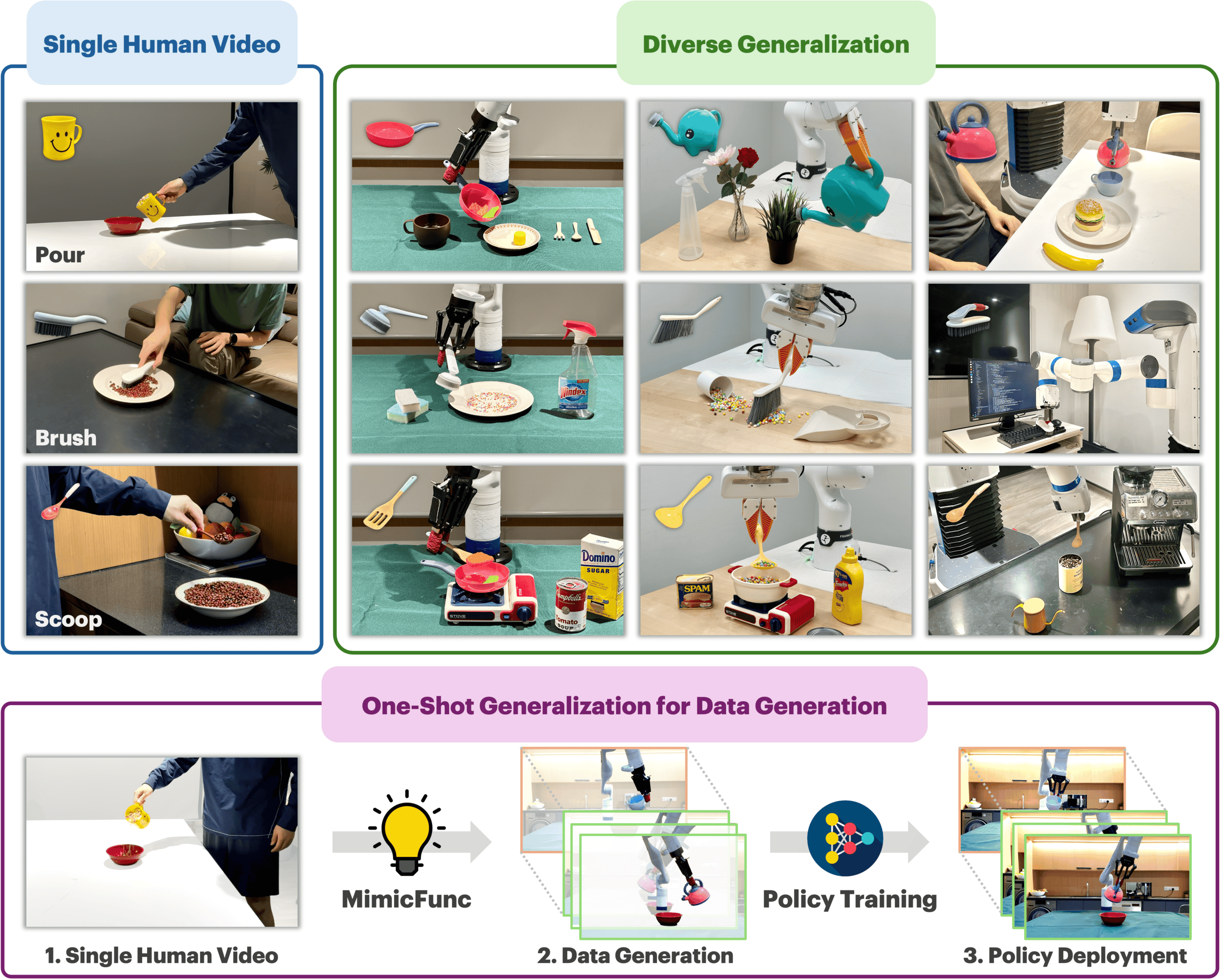}
\vspace{-0.2in}
\caption{Given a single human video, MimicFunc enables the robot to manipulate novel tools for functionally equivalent tasks. Through the one-shot generalization capability, the rollout data generated by MimicFunc can be further leveraged to train visuomotor policies efficiently.}
\vspace{-0.35in}
\label{fig:concept}
\end{figure}

\section{Introduction}

The ability to use tools has long been recognized as a hallmark of human intelligence. While recent advances in developing generalist robots and large robotics foundation models \cite{o2024open, cheang2025gr} have shown promise in endowing robots with similar capabilities, these approaches typically rely on extensive domain expertise and labor-intensive teleoperation data collection. In contrast, humans can mimic tool manipulation behavior by observing others perform a task just once and seamlessly transfer the skill to diverse tools for functionally equivalent tasks. Endowing robots with such an ability not only offers an intuitive approach to teaching robots through human demonstration but also provides a promising and scalable alternative to labor-intensive teleoperation data collection, unlocking the potential to leverage Internet-scale human videos for training visuomotor policies efficiently.

Toward this goal, this paper tackles the problem of imitation of tool manipulation through a single demonstration. The objective is to \textit{enable the robot to imitate from a single human video and generalize the skill to manipulate novel tools for functionally equivalent tasks.} While humans can effortlessly achieve this goal, it remains a non-trivial challenge for robots due to intra-function variations among functionally similar tools, such as differences in shape, size, and topology, as illustrated in Figure~\ref{fig:concept}. The key challenge lies in establishing function-level correspondences among tools in the presence of such variations, which requires capturing invariances in both functionality, understanding how to enable intended uses of tools, and actionability, understanding how to interact with tools to accomplish tasks. Previous methods \cite{vitiello2023one, heppert2024ditto, di2024dinobot, zhu2024vision, li2024okami, biza2023one, zhang2024one, zhudensematcher} typically establish correspondences based on geometric or visual similarities. As a result, they have shown limited flexibility and adaptability.

This limitation motivates us to ask: \textit{How to capture invariances in tool manipulation despite significant intra-function variations?} Pioneering studies in cognitive anthropology \cite{washburn1960tools} reveal that humans exhibit consistent behavioral patterns when using different tools to achieve the same function. For instance, the behavioral pattern of pouring involves approaching the tool, grasping it, and directing its spout toward the target container. This spatiotemporal pattern remains invariant across different tools affording pouring. Inspired by this observation, we propose MimicFunc, which emphasizes the functional aspects of correspondences over geometric or visual similarities for tool manipulation imitation. MimicFunc achieves this by establishing \textit{functional correspondences} with function frame, a function-centric local coordinate frame constructed with keypoint-based abstraction, consisting of a function point, capturing tool–target interaction and serving as a spatial anchor for function frame construction; a grasp point, capturing hand–tool interaction; and a center point, serving as a consistent, object-agnostic reference for defining the function frame. Such a state representation captures the invariant spatiotemporal pattern of tool manipulation, while ignoring function-irrelevant geometric details, enabling consistent and interpretable correspondences across tools for one-shot, generalizable skill transfer.

Technically, MimicFunc is factorized into three stages: (1) Functional keypoint extraction, which detects 3D functional keypoints and extracts their motions from the human video; (2) Functional correspondence establishment, which constructs function frames with 3D functional keypoints and establishes functional correspondences via function frame alignment; (3) Function frame-based action generation, which transfers the skill by synthesizing a motion trajectory through function frame-based optimization for robot execution. Extensive real-robot experiments on diverse tool manipulation tasks demonstrate that, given a single RGB-D human video, MimicFunc enables superior generalization to novel tools with significant intra-function variations, as well as to novel spatial configurations, different embodiments, and environments. Furthermore, leveraging MimicFunc's one-shot generalization capability, the generated rollouts can be used to train visuomotor policies without requiring labor-intensive teleoperation data collection for novel objects.

\section{Related Work}
\textbf{Imitation-based Robotic Manipulation.} Imitating human behavior has been a long-standing approach in robotic manipulation \cite{argall2009survey, ravichandar2020recent}. Recently, there has been a growing trend of training visuomotor policies using BC frameworks \cite{chi2023diffusion, zhao2023learning, ze20243d} on expert demonstrations. However, these approaches typically rely on substantial domain expertise and labor-intensive teleoperation data collection. Another line of research explores one-shot imitation learning \cite{zhang2024one, finn2017one, duan2017one, yu2018one}, yet these approaches often require extensive data collection for meta-training and struggle to generalize to out-of-domain objects with geometric or visual differences. More closely related to our work, recent studies \cite{vitiello2023one, heppert2024ditto, di2024dinobot, zhu2024vision, li2024okami, biza2023one} investigate imitating tool manipulation from a single human video through techniques such as keypoint-based pose estimation \cite{vitiello2023one, heppert2024ditto, di2024dinobot}, global point set registration \cite{zhu2024vision, li2024okami}, and shape warping \cite{biza2023one}. Nevertheless, these methods typically assume that demo and test tools share highly similar shapes or appearances, which limits their generalization to novel tools. More recently, DenseMatcher ~\cite{zhudensematcher} learns 3D dense correspondences to enable category-level manipulation from a single demonstration. However, it exhibits poor generalization under large cross-category, intra-function variations. Our work is also closely related to motion retargeting \cite{li2024okami, gleicher1998retargetting, hu2014online}, as both aim to transfer motions across embodiments with differing kinematics. In fact, MimicFunc can be viewed as a specialized form of motion retargeting that uniquely focuses on object-centric functional intent rather than purely on embodiment-level motion transfer.

\textbf{Keypoint Representation for Tool Manipulation.} Keypoint representation has been extensively studied in tool manipulation \cite{manuelli2019kpam, liu2024moka, huangrekep, gao2023k, gao2024bi}. For instance, KPAM \cite{manuelli2019kpam} and K-VIL \cite{gao2023k, gao2024bi} leverage 3D semantic keypoint representation to accomplish category-level tool manipulation tasks. More recent works leverage foundation models to predict semantic keypoints for tool manipulation. MOKA \cite{liu2024moka} generates planar manipulation motions via mark-based visual prompting \cite{nasirianypivot}, while MimicFunc extends this idea by synthesizing 3D tool manipulation trajectories, supporting more complex tasks. ReKep \cite{huangrekep} encodes manipulation tasks as task-specific keypoint constraints in VLM prompts, but these require substantial manual effort and hand-engineering. In contrast, MimicFunc automatically extracts constraints from human videos and enables more effective trajectory generation. More importantly, unlike previous methods ~\cite{liu2024moka, huangrekep, gao2023k, gao2024bi} that extract keypoints without explicit semantic grounding, MimicFunc builds upon an abstraction that captures both functional and physical semantics, forming a structured ``functional skeleton" of the tool. Such a formulation enables more consistent and interpretable correspondences across tools for function-level generalization.

\begin{figure}[t]
\centering
  \vspace*{-0.4in}
\includegraphics[width=1\linewidth]{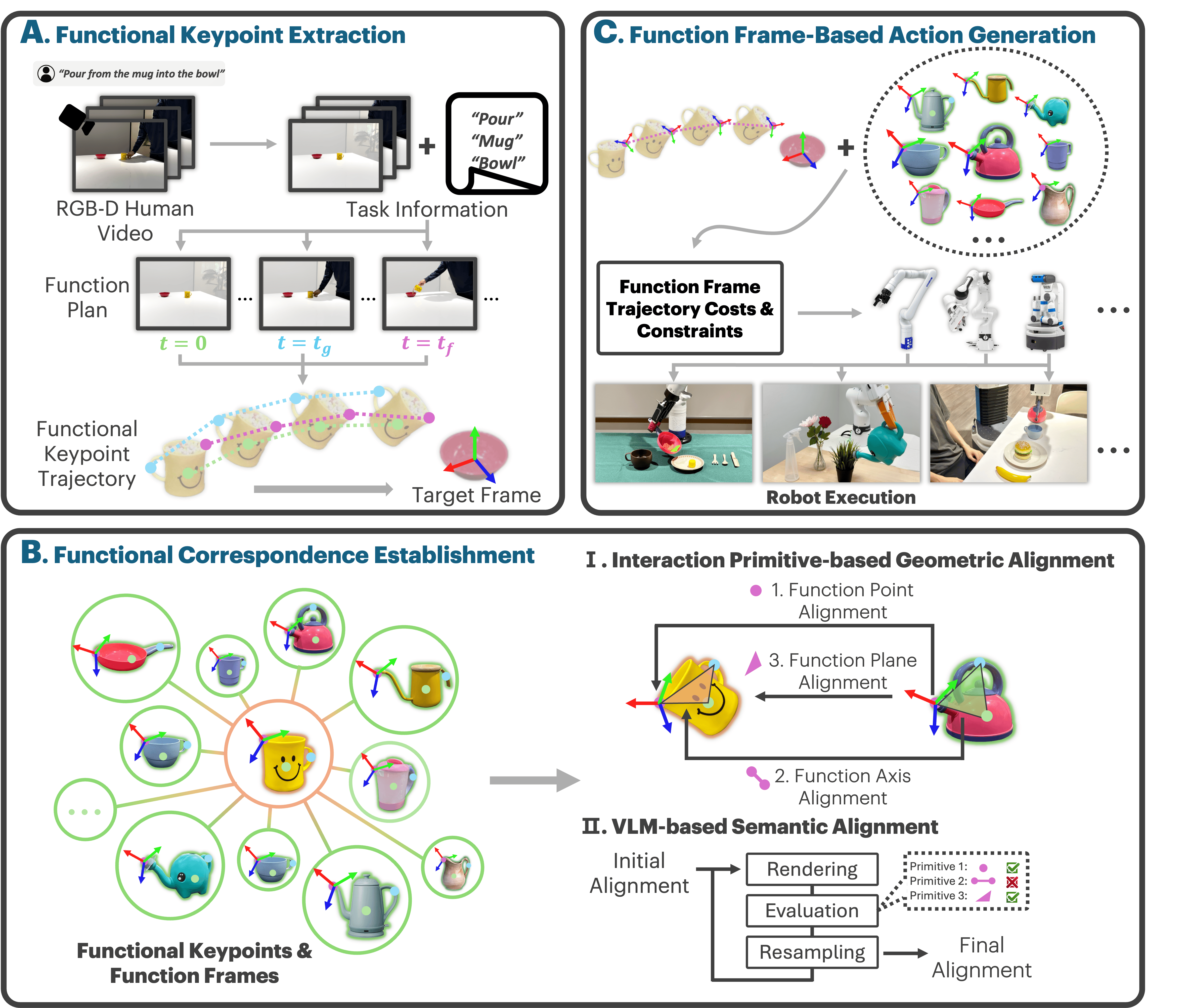}
\vspace{-0.2in}
\caption{\textbf{Overview of MimicFunc Pipeline.} MimicFunc consists of three stages: (1) Functional keypoint extraction from human video, (2) Functional correspondence establishment with function frame, and (3) Function frame-based action generation.}
\vspace{-0.2in}
\label{fig:pipeline}
\end{figure}

\section{MimicFunc}

In this section, we introduce MimicFunc, a method for imitating tool manipulation from a single human video. MimicFunc consists of three stages: (1) Functional keypoint extraction from human video  (Section \ref{detection}), (2) Functional correspondence establishment with function frame (Section \ref{correspondence}), and (3) Function frame-based action generation (Section \ref{planning}). Each will be detailed for the rest of this section. An overview of the pipeline is presented in Figure~\ref{fig:pipeline}. 

\subsection{Problem Formulation}
We consider the problem of enabling the robot to imitate the tool manipulation behavior from a single RGB-D human video to accomplish functionally equivalent tasks using novel tools. Specifically, each task involves manipulating a tool (object) to interact with a target (object) in a tabletop environment. During the demonstration phase, a human performs a tool manipulation task, recording a sequence of RGB-D frames, $\mathcal{V}_H = \{I_t\}_{t=0}^{N-1}$, where $N$ denotes a finite task horizon. The sequence $\mathcal{V}_H$ is paired with a natural language task description $l_H$ (e.g., \textit{``use the \underline{mug} to \underline{pour} contents into the \underline{bowl}"}) that specifies three elements: a tool, a target, and a function. During inference, with novel tools,  environments, and task configurations, the objective is to map the robot observation $o_R$ and task description $l_R$ to a trajectory $\tau_R = \{a_t\}_{t=0}^{N-1}$ for robot execution. Here, $a_t = (R_t, T_t) \in \text{SE(3)}$ represents the 6-DoF end-effector pose at timestep $t$, where $R_t \in \text{SO(3)}$ and $T_t \in \mathbb{R}^3$ denote 3D orientation and translation, respectively. 


\subsection{Functional Keypoint Extraction from Human Video}\label{detection}

\noindent \textbf{Function Plan Generation with Keyframe Discovery.} Since a tool manipulation task involves multiple stages, MimicFunc first generates a function plan with keyframe discovery to guide the robot's execution. Specifically, a function plan includes three stages: (1) the initial keyframe $I_0$ ($t=0$), where the tool and target are in their initial states; (2) the grasping keyframe $I_g$ ($t=t_g$), where the hand grasps the tool; and (3) the function keyframe $I_f$ ($t=t_f$), where the tool interacts with the target. These keyframes satisfy the temporal constraint $0 < t_g <  t_f < N-1$. We use VideoCLIP \cite{xu2021videoclip} to discover these keyframes by computing similarity between video frames and predefined keyframe descriptions. For long-horizon tasks, MimicFunc generates a high-level task plan by chaining multiple function plans.

\noindent \textbf{Functional Keypoint Extraction.} MimicFunc detects 3D functional keypoints, representing a structured functional abstraction of the tool, using keyframes from the function plan and tracks their motions. For grasp point detection, MimicFunc first uses HaMeR \cite{pavlakos2024reconstructing} to reconstruct the hand mesh using $I_g$. The grasp point is then determined as the center of the intersection between the fingertip region and the tool. The center point is defined as the 3D bounding box center of the tool in $I_0$. Detecting the function point is non-trivial, as the tool may not physically contact the target (e.g., pouring) and requires commonsense knowledge about tool usage. To address this, MimicFunc employs mark-based visual prompting \cite{nasirianypivot} to identify the function point on $I_f$ (similar to \cite{liu2024moka}) and then projects the point into 3D space. To extract their motions, MimicFunc first estimates the tool's relative transformations between consecutive timesteps using CoTracker \cite{karaev2025cotracker} and then computes the 3D functional keypoint trajectory $\{K_H^t\}_{t=0}^{N-1} = \{[p_{\text{func}}^t, p_{\text{grasp}}^t, p_{\text{center}}^t]\}_{t=0}^{N-1}$, where $p \in \mathbb{R}^3$ and $H$ denotes human. To ensure that the extracted motion is independent of the absolute positions of the tool and target, MimicFunc transforms 3D elements from the camera frame to the target (object) frame by estimating the target object's pose relative to the camera. Unless otherwise specified, all 3D elements are represented in this relative target frame.

\subsection{Functional Correspondence Establishment with Function Frame}\label{correspondence}
\noindent \textbf{Functional Keypoint Transfer.} To construct function frames, MimicFunc first transfers keypoints from the demonstration tool to the test tool in a coarse-to-fine manner. It first performs in-context visual prompting to propose coarse regions for both function and grasp points, using \( p_{\text{func}}^0 \) and \( p_{\text{grasp}}^0 \) as references.  The decision to propose regions rather than directly predict points stems from two factors: (1) VLMs lack point-level correspondence, and (2) they may produce discrete point predictions that do not align with the ideal keypoint locations. Then, a dense semantic correspondence model \cite{zhang2024tale} with learned geometric priors is used to accurately transfer the keypoints to their corresponding regions, resulting in \( q_{\text{func}}^0 \) and \( q_{\text{grasp}}^0 \). Similarly, the test tool can be functionally abstracted as \( K_R^0 = [q_{\text{func}}^0, q_{\text{grasp}}^0, q_{\text{center}}^0] \), where \( q \in \mathbb{R}^3 \) and the \( R \) denotes the robot.

\noindent \textbf{Function Frame Construction.} Based on two sets of keypoint abstractions, MimicFunc constructs function frames to represent function-centric spatiotemporal patterns of tool manipulation. Specifically, given \( K_H^{t} \) and \( K_R^t \), the function frames \( \Pi_H^{t} \) and \( \Pi_R^t \) are constructed as follows.  For \( \Pi_H^{t} \), the origin is placed at the function point \( p_{\text{func}}^{t} \), where the interaction between the tool and target occurs. The orientation is defined by an orthonormal basis constructed from \( K_H^{t} \), where the unit vector from the center to the function point, \( \mathbf{v}_H^{t} \), serves as the principal axis, referred to as the function axis, which provides a stable, reproducible directional cue reflecting how the tool operates. The same procedure is repeated to construct \( \Pi_R^t \), with \( q_{\text{func}}^t \) as the origin and the function axis \( \mathbf{v}_R^{t} \) as the principal axis. More details on function frame construction are available in Appendix \ref{sec:appendix_b}.

\noindent \textbf{Function Frame Alignment.} To transfer functional intent across different tools, MimicFunc aligns the spatiotemporal patterns encoded in function frames. In this section, we specifically focus on function keyframe alignment $\mathbf{\Pi}_{\text{func}}$, which defines the desired test tool state at timestep \( t_f \), denoted as $\mathbf{\Pi}_R^{t_f}$. Aligning this critical state preserves the core functionality of human behavior, while the remaining motion can be flexibly optimized based on task context, as discussed in the following section. Function frame alignment is divided into two stages: (1) an initial stage for interaction primitive-based geometric alignment and (2) a refinement stage for VLM-based semantic alignment.

Inspired by \cite{manuelli2019kpam, gao2023k}, MimicFunc first establishes a coarse alignment by enforcing function-relevant geometric constraints on function frames. These constraints operate on three types of interaction primitives, each corresponding to a physically meaningful spatial element critical for manipulation:
\begin{enumerate}[label=\arabic*., leftmargin=*, itemsep=-2pt, topsep=-2pt]
    \item \textbf{Interaction primitive 1: point.} Function point alignment ensures that the interaction occurs at the intended location on the test tool by aligning $q_{\text{func}}^{0}$ with $p_{\text{func}}^{t_f}$.
    \item \textbf{Interaction primitive 2: axis.} Function axis alignment ensures the proper operational direction for executing function-specific actions (e.g., tilting for pouring) by aligning \( \mathbf{v}_R^0 \) with \( \mathbf{v}_H^{t_f} \).
    \item \textbf{Interaction primitive 3: plane.} Function plane alignment preserves orientation by aligning the normal vectors of \( {\Pi}_R^0 \) and \( {\Pi}_H^{t_f} \).
\end{enumerate}
Each constraint is represented by a $\text{SE(3)}$ transformation. Despite satisfying the geometric constraints derived above, the resulting interaction may still be functionally invalid, primarily due to (1) inaccurate perception and (2) structural differences between tools. To improve the robustness and adaptability, we further incorporate a VLM-based state evaluator for semantic refinement, similar to \cite{pan2025omnimanip}. Specifically, MimicFunc first renders the predicted function keyframe interaction by back-projecting the combined point cloud of the test tool and the target onto the camera plane. It then prompts the VLM to evaluate whether the predicted state is functionally valid. If deemed valid, the alignment is accepted for action generation. Otherwise, the VLM sequentially checks each primitive to automatically identify those contributing to failure. Guided by this feedback, MimicFunc uniformly resamples candidate (points or axes) around the initial constraint and iteratively repeats the process until a valid alignment is found. The alignment process is illustrated in Figure \ref{fig:pipeline}(b).

\subsection{Function Frame-Based Action Generation}\label{planning}
To enable a functionally consistent rollout of the intended behavior, MimicFunc computes the complete function frame trajectory for the test tool, formulated as a constrained optimization problem, using the human demonstration as the reference:
\begin{align*}
\min_{\{\mathbf{\Pi}_R^t\}_{t=0}^{N-1}} \sum_{t=0}^{N-1} \left( \| q_{\text{func}}^t - p_{\text{func}}^t \|_2^2 + \| \text{Log}(\mathbf{R}_R^t (\mathbf{R}_H^t)^\top) \|^2 \right) \quad \text{s.t.} \quad \mathbf{\Pi}_R^0 = \mathbf{\Pi}_{\text{init}}, \; \mathbf{\Pi}_R^{t_f} = \mathbf{\Pi}_{\text{func}}
\end{align*}
where $\text{Log}:\text{SO(3)} \rightarrow \mathbb{R}^3$~\cite{sola2018micro}, and $\mathbf{R}^t$ denotes the rotation matrix derived from the function frame. The constraints $\mathbf{\Pi}_{\text{init}}$ and $\mathbf{\Pi}_{\text{func}}$ represent the initial and function keyframe alignments, respectively.  This optimization framework provides a flexible mechanism for enforcing semantic-geometric alignment between function frames and can be extended to incorporate additional terms such as trajectory smoothness and collision avoidance. Implementation details are provided in Appendix \ref{sec:appendix_c}.

For robot execution, the test function frame trajectory is first transformed into the robot base frame. Then, MimicFunc samples a 6-DoF grasp pose around $q_{\text{grasp}}^0$ on the test tool. The robot end-effector trajectory $\tau_R$ is subsequently computed and executed.

\section{Experiments}
The experimental section aims to answer the following questions: (1) How well does MimicFunc generalize from a single human video to novel tools? (2) How does MimicFunc perform compared to existing methods? (3) How does MimicFunc perform in long-horizon tasks? (4) Can the rollout trajectories generated by MimicFunc be leveraged to train visuomotor policies?

\subsection{Experimental Setup}

\noindent \textbf{Baselines.} We compare MimicFunc against the following baselines: (1) \textbf{\textsc{DINOBot}} \cite{di2024dinobot}, which uses DINOv2 \cite{oquab2024dinov2} to perform semantic feature extraction and correspondence. (2) \textbf{\textsc{DITTO}} \cite{heppert2024ditto}, which employs LOFTR \cite{sun2021loftr} for local feature matching. (3) \textbf{\textsc{ORION}} \cite{zhu2024vision}, which establishes geometric correspondences with point cloud global registration. We adopt the original correspondence implementations of these baselines while keeping the low-level execution consistent with MimicFunc.

\noindent \textbf{Task Description.} We evaluate each method on five functions: \texttt{Pour}, \texttt{Cut}, \texttt{Scoop}, \texttt{Brush}, and \texttt{Pound}. A task is defined by pairing a function with a tool and a target. For each function, we design five tasks using different tools, divided into three levels of generalization: (1) spatial generalization, (2) instance generalization, and (3) category generalization. 

\noindent \textbf{Experimental Protocol.} Each method is evaluated on 25 tasks across five functions, with 10 trials per task. The detailed task success conditions are described in Appendix \ref{sec:appendix_a}. The average success rate is used as the evaluation metric. Test objects are randomly initialized within the intersection of the camera view and the robot workspace.

\subsection{Experimental Results}
\begin{figure}[t]
\centering
  \vspace*{-0.4in}
\includegraphics[width=1\linewidth]{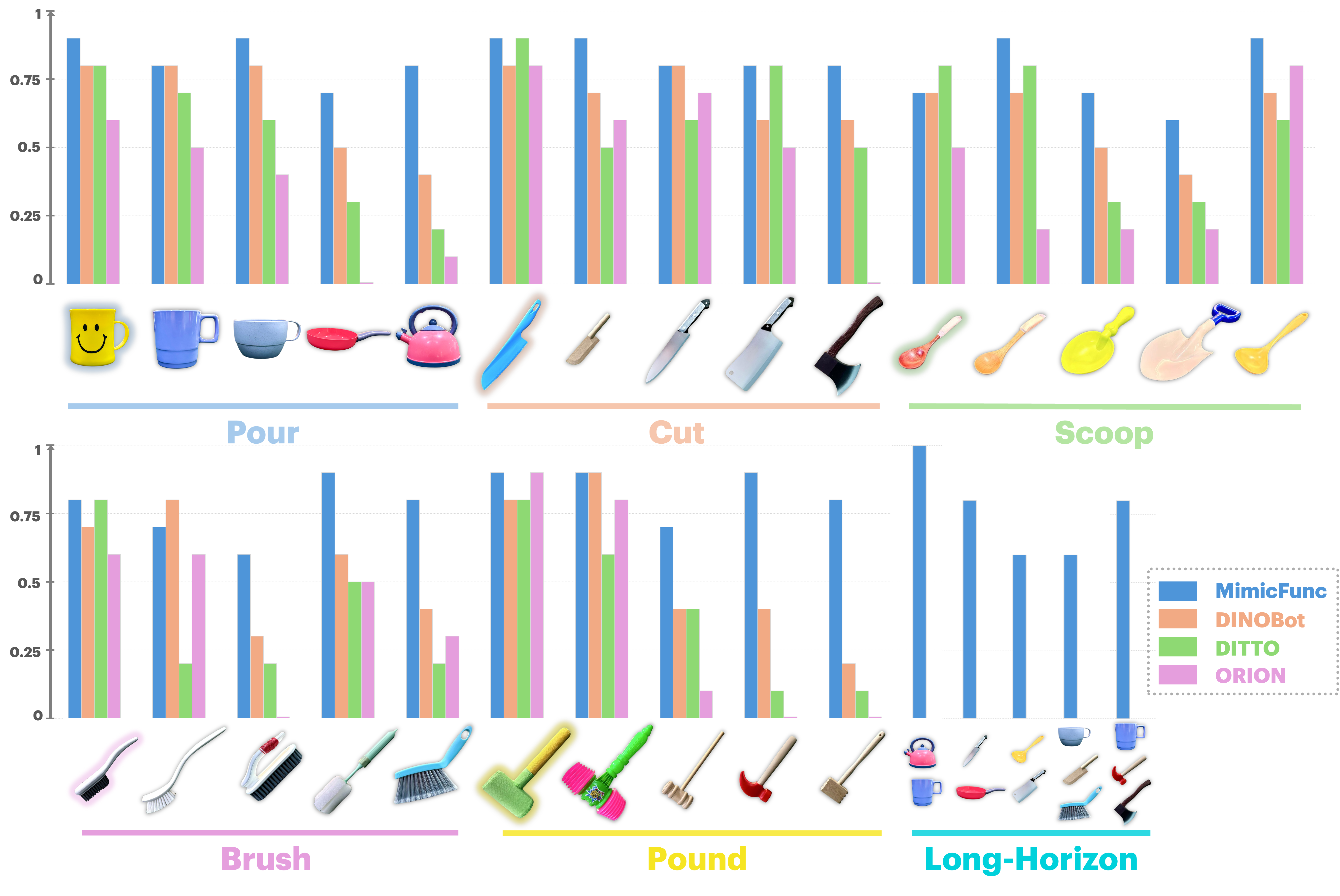}
\vspace{-0.3in}
\caption{Quantitative comparison to baselines. Highlighted tools are used in human videos.}
\vspace{-0.25in}
\label{fig:main_result}
\end{figure}

\begin{figure}[t]
\centering
  \vspace*{-0.4in}
\includegraphics[width=1\linewidth]{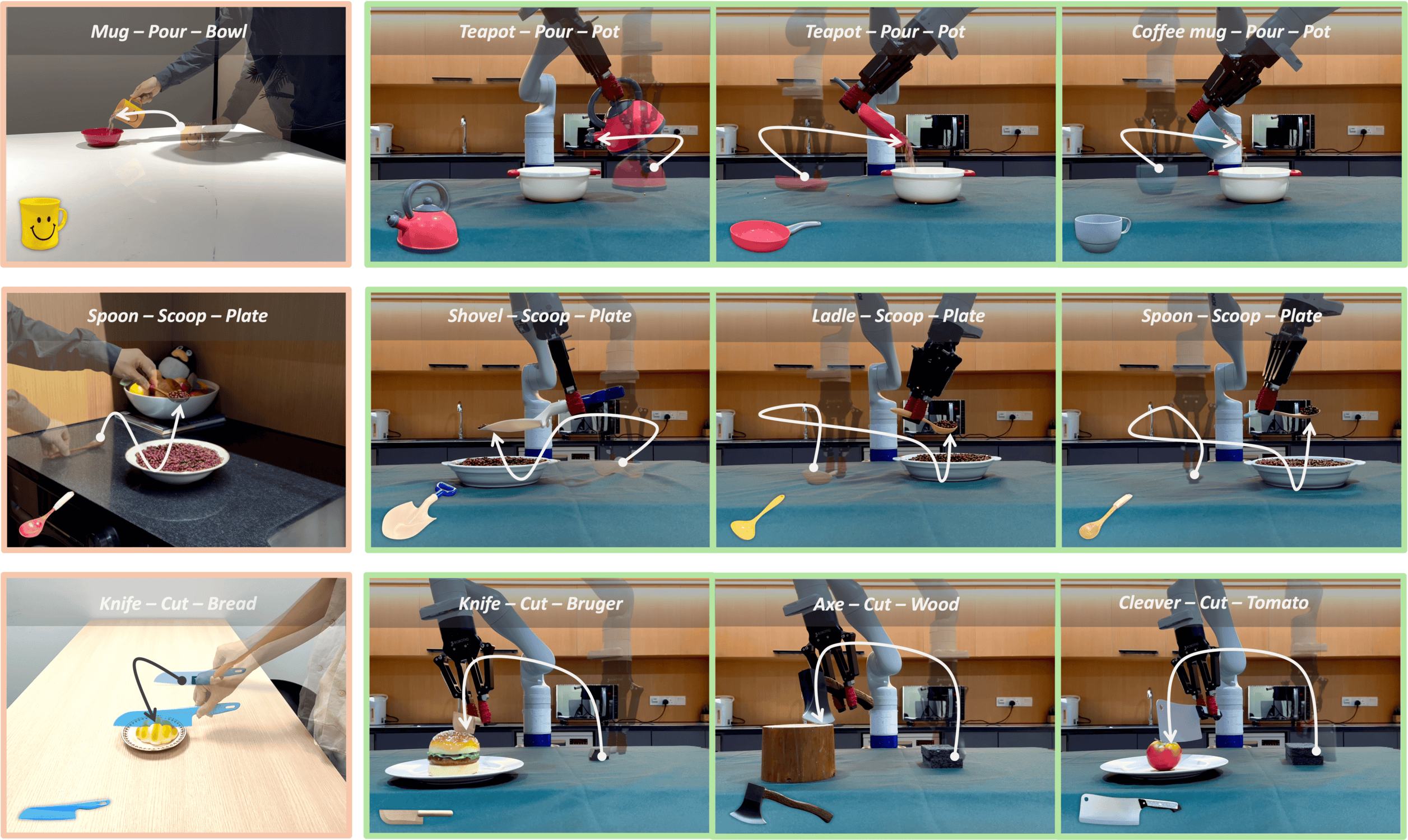}
\vspace{-0.2in}
\caption{Visualization of grasping and function keyframes of human demonstrations and robot rollouts for \texttt{Pour}, \texttt{Scoop}, and \texttt{Cut}.}
\vspace{-0.25in}
\label{fig:exp_qualitative}
\end{figure}

\noindent \textbf{Quantitative Comparison to Baselines.} The quantitative results are reported in Figure~\ref{fig:main_result}. For each function, the first object is used to evaluate spatial generalization, the next two evaluate instance-level generalization, and the final two evaluate category-level generalization. All methods perform reasonably well in spatial generalization, achieving success rates above 70\%. However, all baselines exhibit significant performance drops (from 20\% to 40\%) when generalizing to novel tool instances and categories, especially for those with substantial intra-function variations. Among all baselines, ORION relies solely on geometric features, rendering it ineffective at handling large intra-function variations. DINOBot outperforms both DITTO and ORION, achieving an average success rate of 57.5\% when generalizing to novel tools. This performance can be attributed to DINO's strong visual correspondence capability. However, DINOBot still struggles to establish correspondences between visually distinct tools. In contrast, MimicFunc significantly outperforms all baselines, achieving a high success rate of 79.5\% across five functions for novel tool generalization. Figure~\ref{fig:exp_qualitative} visualizes the qualitative results of real-robot executions.

\begin{wraptable}{r}{0.55\textwidth} 
\centering
\renewcommand\arraystretch{1.2}
\setlength\tabcolsep{4pt}
\begin{tabular}{l ccccc|c}
\toprule
\textbf{Task} & P-P  & C-P  & S-C & P-C-B & P-O-C & \textbf{Overall} \\ \hline
\textbf{TS} & 5/5 & 4/5 & 3/5 & 3/5 & 4/5 & \textbf{76.0\%} \\
\textbf{SC} & 10/10 & 8/10 & 7/10 & 10/15 & 13/15 & \textbf{80.0\%} \\ 
\bottomrule
\end{tabular}
\caption{Quantitative results on long-horizon tasks. Abbreviations: P = \texttt{Pour}, C = \texttt{Cut}, S = \texttt{Scoop}, B = \texttt{Brush}, O = \texttt{Pound}.}
  \vspace{-0.15in} 
\label{fig:multi_tool_exp}
\end{wraptable}

\noindent \textbf{Evaluation on Long-Horizon Tasks.} We evaluate the performance of MimicFunc on long-horizon tasks involving multiple sequential steps. MimicFunc first generates a high-level task plan by chaining multiple function plans and then executes them sequentially. As reported in Table~\ref{fig:multi_tool_exp}, MimicFunc achieves a 76.0\% task success (TS) rate, slightly lower than the single-step setting, and an 80\% step completion (SC) rate. The primary challenge arises from the limited reachability of the single-arm manipulator in larger object layouts. Nonetheless, by explicitly representing skills as function frame trajectories, MimicFunc can be integrated with Task and Motion Planning frameworks \cite{garrett2020pddlstream} to handle more complex long-horizon tasks with geometric constraints.

\begin{wrapfigure}{r}{0.5\textwidth}  
    \centering
    \includegraphics[width=0.5\textwidth]{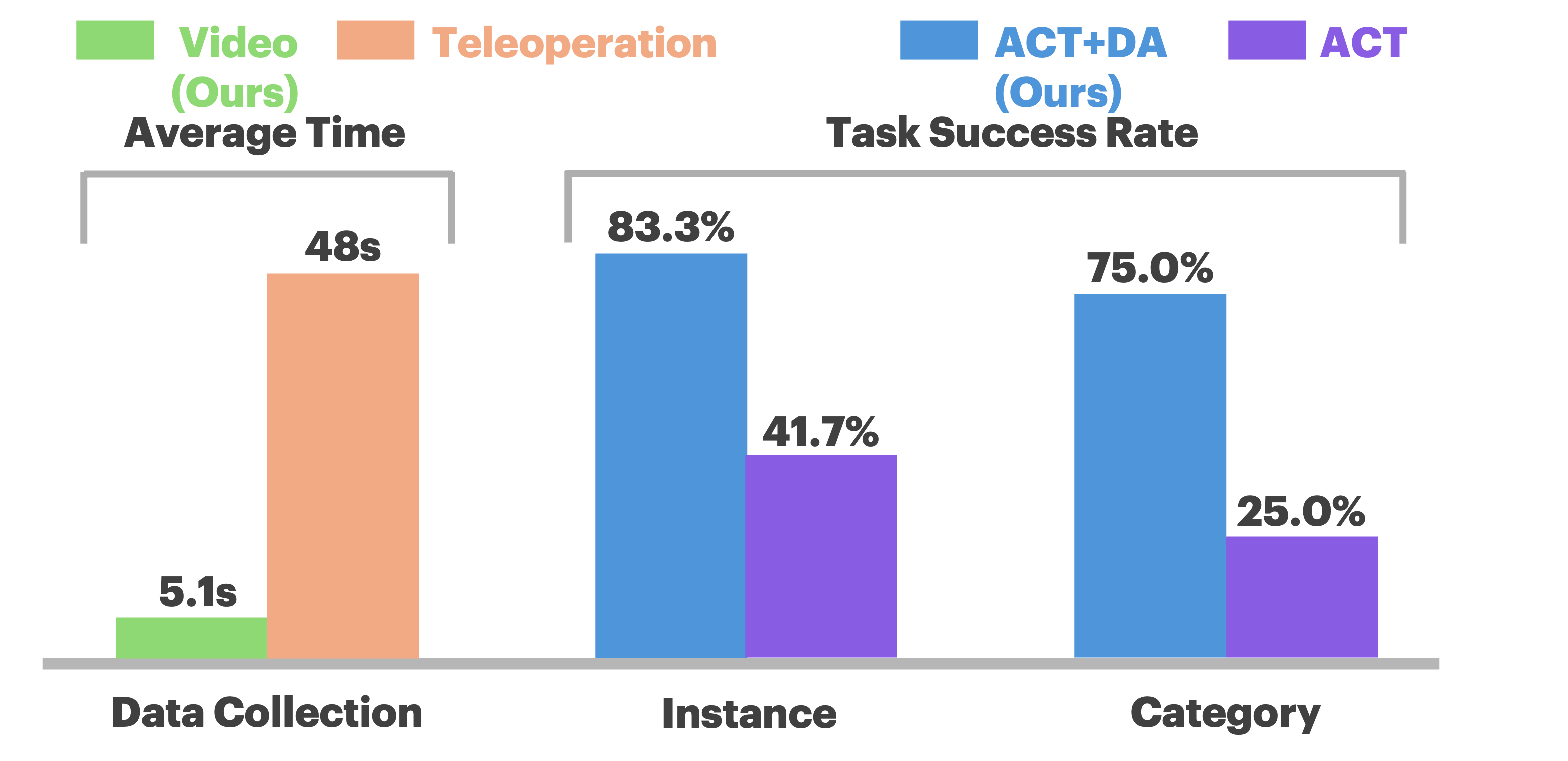}
    \vspace{-0.2in}
    \caption{Performance evaluation of visuomotor policy training.}
    \label{fig:data_exp_result}
     \vspace{-0.1in}
\end{wrapfigure}

\noindent \textbf{Evaluation on Data Generation for Visuomotor Policy Training.} To support the claim that the rollout data generated by MimicFunc can be leveraged for visuomotor policy training, we conduct experiments using the BC method ACT~\cite{zhao2023learning} on \texttt{Pour}. We first collect 50 teleoperation demonstrations to train ACT. As shown in Figure~\ref{fig:data_exp_result}, ACT exhibits limited generalization to novel instances and categories. To enhance its performance, we deploy MimicFunc on these novel objects over randomly initialized layouts, automatically generating 30 rollouts per object using only a single human video. Successful trajectories are collected to train ACT. As illustrated in Figure~\ref{fig:data_exp_result}, ACT+DA (data augmentation)  improves performance on these instances and categories by 41.6\% and 50.0\%, respectively, using high-quality, consistently generated data from MimicFunc without requiring labor-intensive teleoperation data collection for novel objects. Moreover, this approach surpasses the variability and limited precision typically observed in human teleoperation. In terms of data collection efficiency, each teleoperation demonstration takes approximately 48 seconds, while MimicFunc requires only 5.1 seconds on average to capture a human video. These results demonstrate the potential of MimicFunc as a scalable and efficient data generator for visuomotor policy learning.

\begin{wrapfigure}{r}{0.5\textwidth} 
    \centering
    \includegraphics[width=0.48\textwidth]{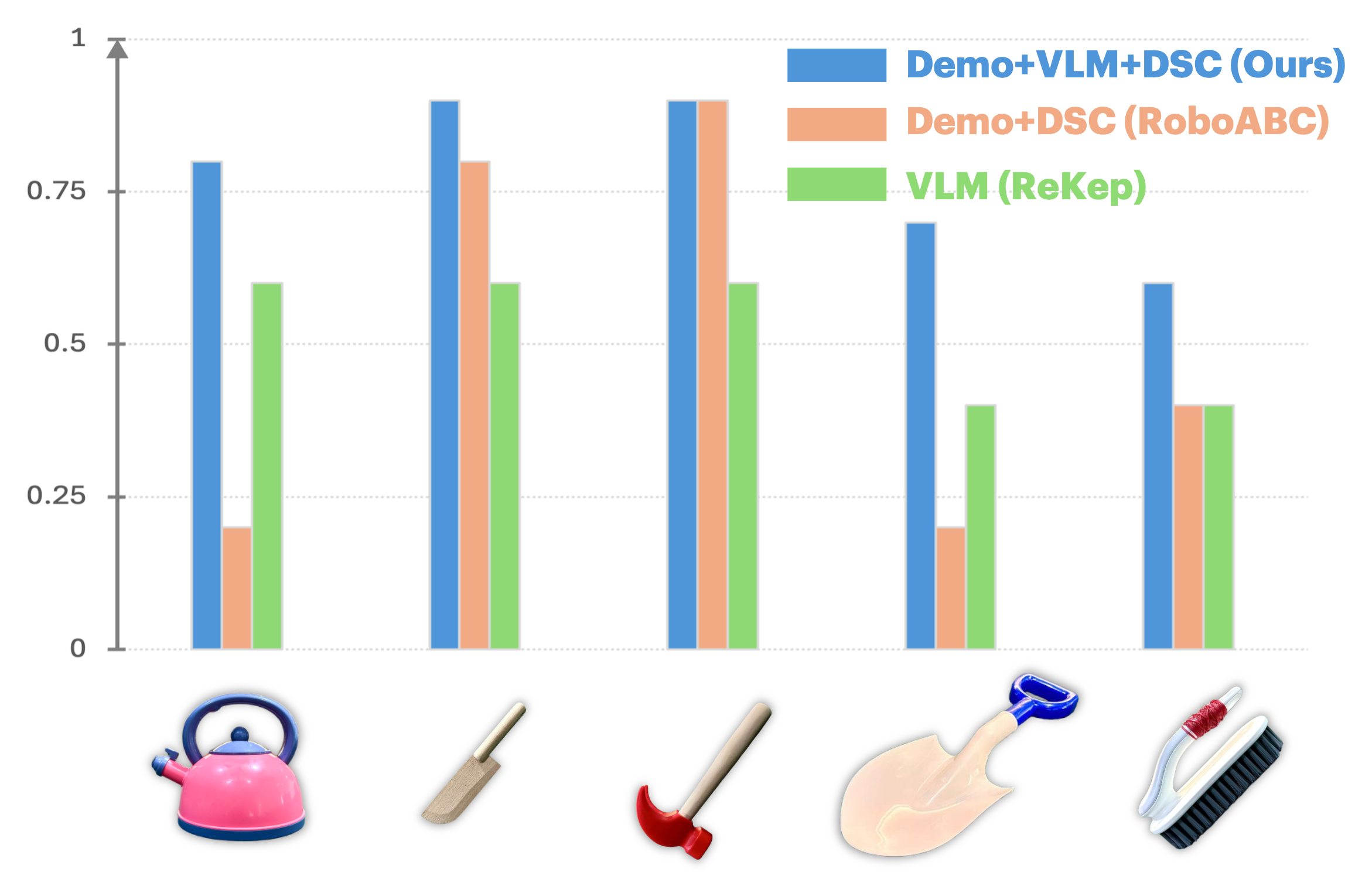}
    \vspace{-0.1in}
    \caption{Ablation study results.}
    \label{fig:ablation_result}
     \vspace{-0.1in}
\end{wrapfigure}

\noindent \textbf{Ablation Study.} We conduct an ablation study on the functional keypoint transfer strategy. Three strategies are evaluated: (1) Demo+VLM+DSC (ours); (2) Demo+DSC, which relies solely on a dense semantic correspondence model for keypoint transfer, following the approach in Robo-ABC~\cite{ju2024robo}; and (3) VLM only, which directly prompts the VLM to propose keypoints in a zero-shot manner, as done in ReKep~\cite{huangrekep}. As shown in Figure~\ref{fig:ablation_result}, the proposed strategy consistently outperforms ablated versions, demonstrating that (1) the dense semantic correspondence model alone struggles with large intra-function variations, and (2) incorporating human demonstrations as references for VLMs significantly improves keypoint localization accuracy.

\section{Conclusion}
In this work, we present MimicFunc, a method for imitating tool manipulation from a single human video. At the core of MimicFunc is the ability to establish functional correspondences with function frame. This enables robots to generalize the skill from a single human video to novel tools despite significant intra-function variations. Furthermore, leveraging MimicFunc’s one-shot generalization capability, the generated rollouts can be used to train visuomotor policies without requiring labor-intensive teleoperation data collection for novel objects.

\section{Limitations and Future Work}

Despite the promising results, several limitations remain that point to directions for future work. (1) MimicFunc currently relies on RGB-D input, which limits its applicability in directly leveraging RGB Internet human videos. Extending MimicFunc to support RGB videos could further enhance flexibility and reduce the human effort required for data collection. A potential solution is incorporating monocular depth estimation models, such as Depth Anything, to infer depth information from RGB inputs. (2) Although we have demonstrated the potential of data generation for visuomotor policy training, the current system remains limited in scope. Future work will aim to extend MimicFunc into a complete system in simulated environments capable of efficiently generating diverse data from a single human demonstration for visuomotor policy learning. (3) Complex tool manipulation tasks may involve bimanual coordination, whereas the current implementation only considers single-handed manipulation. Future work will extend MimicFunc to support dual-arm or even multi-fingered coordination to handle more complex manipulation scenarios.

\section{Acknowledgement}
We thank Linfeng Li (NUS) for helpful discussion and technical support. This work was supported in part by the Shenzhen Science and Technology Program (No. SGDX20240115111759002) and in part by Meituan.


\bibliography{example}  

\clearpage

\appendix
\section{Appendix}

\subsection{Experimental Setup and Results} \label{sec:appendix_a}

\subsubsection{Task Success Conditions}

\begin{itemize}[leftmargin=*]
    \item \texttt{Pour}: The particles within the tool are transferred into the target container.
    \item \texttt{Cut}: The blade of the tool makes contact with the target from above.
    \item \texttt{Scoop}: The tool collects and securely holds particles from the target container.
    \item \texttt{Pound}: The bottom of the tool head strikes the nail head.
    \item \texttt{Brush}: The tool moves across the target’s surface, displacing particles with its bristles. \\
\end{itemize}

\subsubsection{Examples of Human Demonstrations}

\begin{figure}[h]
\centering
\includegraphics[width=1\linewidth]{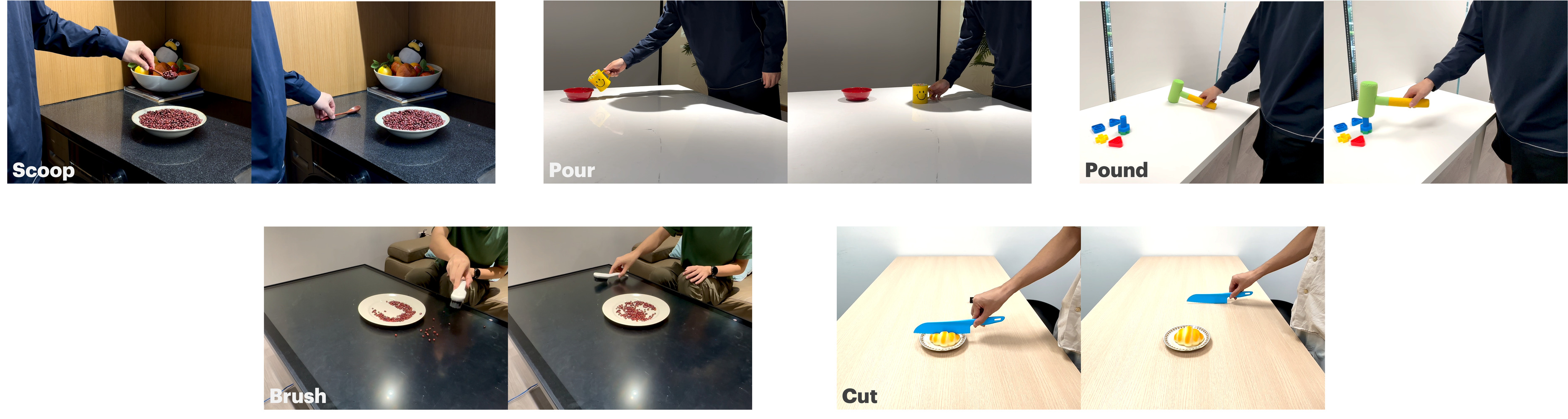}
\vspace{-0.2in}
\caption{Visualization of grasping and function keyframes of human demonstrations.}
\end{figure}

\subsubsection{Failure Analysis}

\begin{wrapfigure}{r}{0.5\textwidth} 
    \centering
    \includegraphics[width=0.48\textwidth]{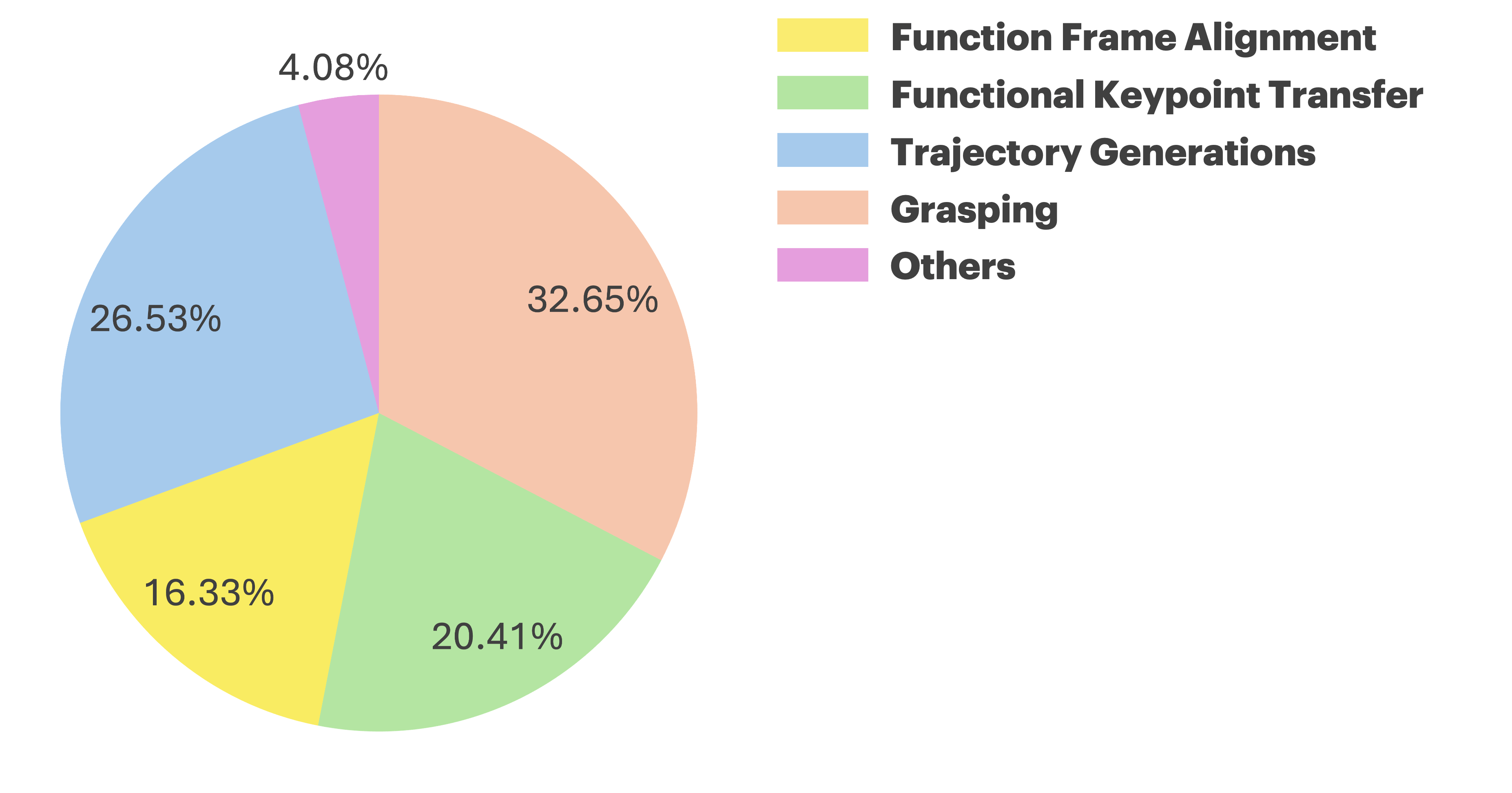}
    \caption{Failure analysis of system components.}
    \label{fig:failure}
     \vspace{-0.1in}
\end{wrapfigure}

The modular design of MimicFunc facilitates the interpretation and in-depth analysis of failure cases. The result of the failure analysis is reported in Figure \ref{fig:failure}. The identified failure sources are categorized into: (1) function frame alignment, (2) functional keypoint transfer, (3) trajectory generation, (4) grasping, and (5) others (e.g., segmentation, detection). 

The primary failures arise from (4) and (3). Grasping failures often arise from the inherent constraints of certain gripper types, especially rigid or underactuated designs. These grippers may lack the adaptability required to conform to diverse object geometries or provide sufficient contact stability, leading to issues such as tool flipping or slipping during tasks. Such failures are primarily due to unstable contact between the tool and the gripper, preventing the robot from completing its intended actions. Failures in trajectory generation primarily result from unexpected contact between the tool and the target, particularly in contact-rich tasks (e.g., ``\textit{use scrubber to brush the plate}"). These tasks require precise force application and adaptability to varying contact conditions. Providing visual-tactile feedback is essential for successfully accomplishing such tasks. Functional keypoint transfer errors are mainly caused by incorrect candidate region proposals for function points, but contribute less significantly to overall failures. These errors may be mitigated as VLMs continue to improve. Function frame alignment errors are mainly attributed to inaccurate depth information of the functional keypoints. Empirically, the functional correspondences are well established with accurate 3D functional keypoint locations. Ensuring precise depth sensing and calibration can significantly reduce these alignment errors.

\subsubsection{Target Frame Detection}
To ensure that the tool motion is independent of the absolute positions of both the tool and the target, MimicFunc transforms all 3D elements from the camera coordinate frame into the target object’s coordinate frame. This transformation requires estimating the target object’s pose relative to the camera, which our method currently performs without relying on pre-existing mesh models. Instead, the target frame is estimated on the fly from the observed scene.

Specifically, we first acquire the target object’s segmented point cloud from the camera. We then compute its principal axes via Principal Component Analysis (PCA) to determine the dominant orientation. The origin of the target frame is set to the center of the object’s 3D bounding box, computed from the point cloud. Among the principal axes, the one most closely aligned with the estimated surface normal is assigned as the z-axis, ensuring that the frame is physically consistent with the object’s geometry. The remaining axes are chosen to form a right-handed coordinate system, preserving orthogonality.

This procedure produces a stable, object-centered reference frame that remains robust to variations in both object position and camera viewpoint. Moreover, it can be seamlessly integrated with existing pose estimators (e.g., FoundationPose) to further improve robustness and adaptability, ensuring consistent target frame detection across diverse tasks, object geometries, and environments.

\clearpage

\subsubsection{Qualitative Results}

\begin{figure}[h]
\centering
\includegraphics[width=1\linewidth]{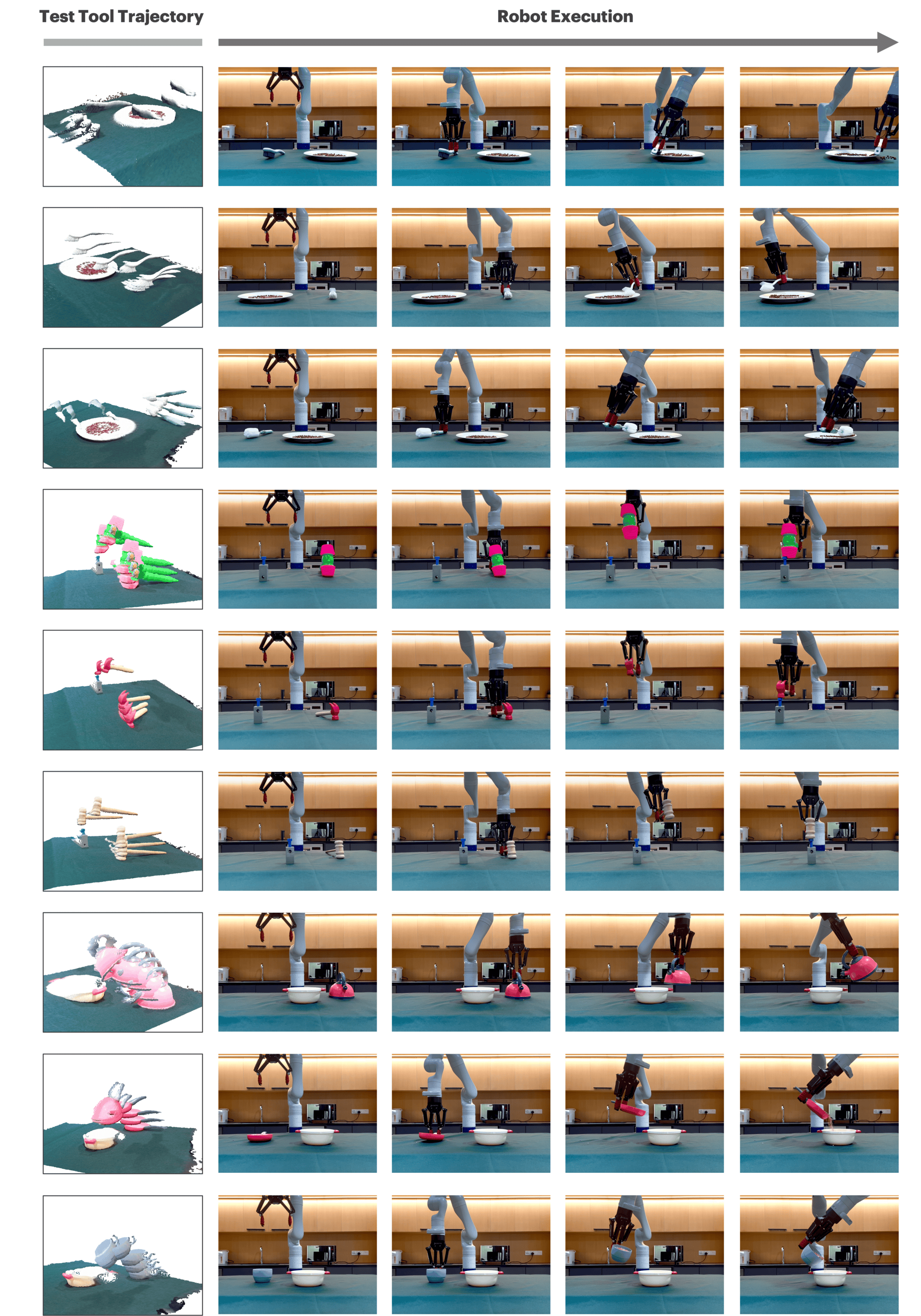}
\end{figure}

\begin{figure}[h]
\centering
\includegraphics[width=1\linewidth]{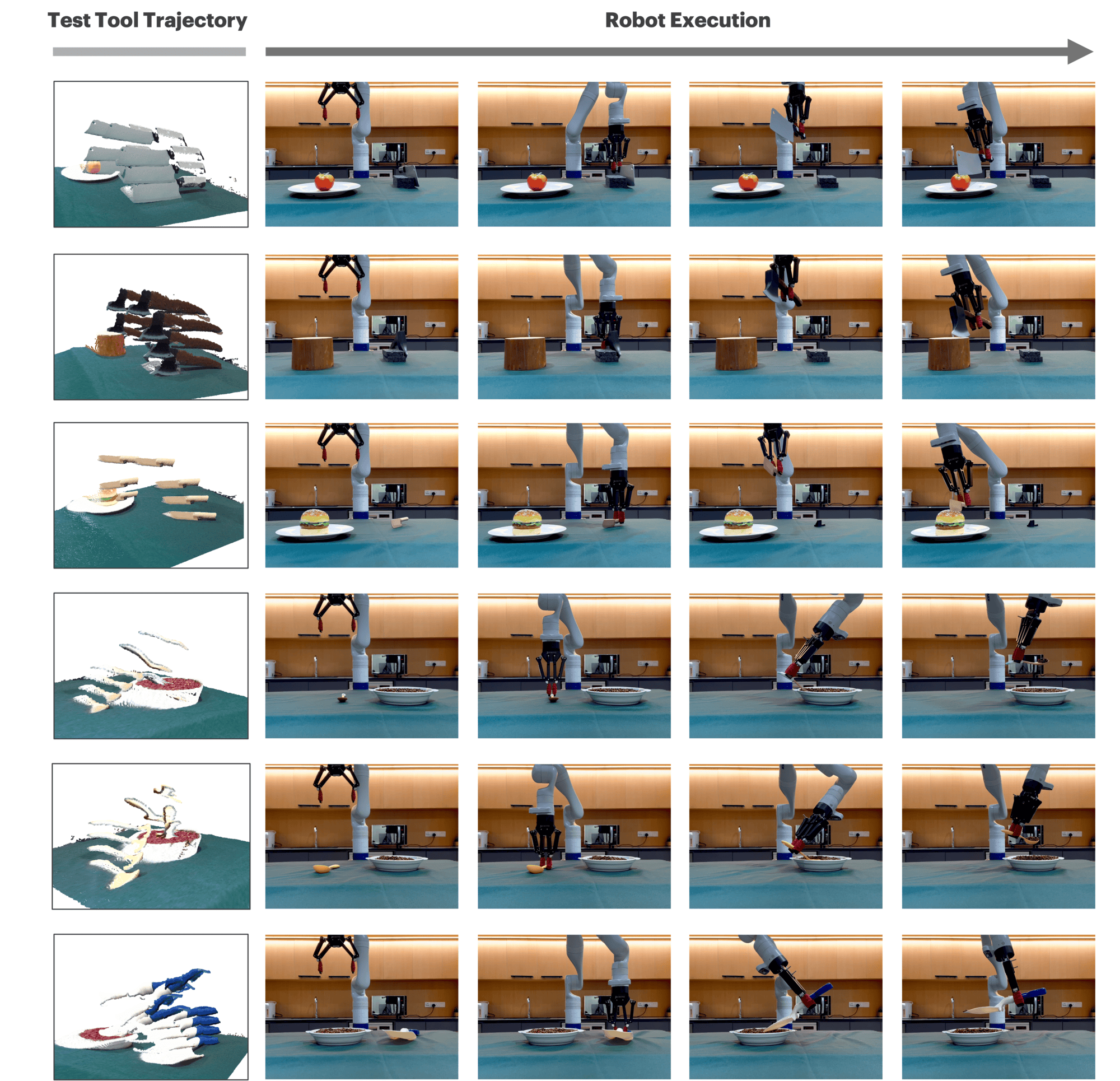}
\caption{Qualitative results of real-robot executions.}
\end{figure}

\clearpage

\subsection{Function Frame Construction and Alignment}\label{sec:appendix_b}

\subsubsection{Function Keypoint Transfer}

The pseudo-code for functional keypoint transfer is illustrated in Algorithm \ref{alg:3d_keypoint_transfer}.

\begin{algorithm}[h]
\caption{Functional Keypoint Transfer.}
\label{alg:3d_keypoint_transfer}
\textbf{Input:} \\
\hspace{1em} Demo functional keypoints \( K_H^0 = [p_{\text{func}}^0, p_{\text{grasp}}^0, p_{\text{center}}^0] \), Initial keyframe \( I_0 \), Robot observation \( o_R \), Test tool mask \( M \), \\
\hspace{1em} Dense semantic correspondence model \( \Phi \), \\
\hspace{1em} 3D-2D projection \( P_{\text{3D-2D}} \), 2D-3D projection \( P_{\text{2D-3D}} \), 3D center computation \( F_{\text{center}} \) \\
\textbf{Output:} Test functional keypoints \( K_R^0 = [q_{\text{func}}^0, q_{\text{grasp}}^0, q_{\text{center}}^0] \)

\begin{algorithmic}[1]
    \State \( K_R \gets \emptyset \)

    \State \textbf{1. Coarse-Grained Region Proposal:}
    \For{each \( k \in \{\text{func}, \text{grasp}\} \)}
        \State \( p_k^{2D} \gets P_{\text{3D-2D}}(p_k^0, I_0) \)
        \State \( r_k \gets \text{VLM}(p_k^{2D}, I_0, o_R, M) \) \Comment{Region proposal}
    \EndFor

    \State \textbf{2. Fine-Grained Point Transfer:}
    \For{each \( k \in \{\text{func}, \text{grasp}\} \)}
        \State \( q_k^{2D} \gets \Phi(p_k^{2D}, r_k, I_0, o_R) \) \Comment{Point transfer}
        \State \( q_k^0 \gets P_{\text{2D-3D}}(q_k^{2D}, o_R) \)
    \EndFor

    \State \textbf{3. 3D Center Computation:}
    \State \( q_{\text{center}}^0 \gets F_{\text{center}}(M, o_R) \)

    \State \textbf{4. Functional Keypoint Transfer Output:}
    \State \( K_R^0 \gets [q_{\text{func}}^0, q_{\text{grasp}}^0, q_{\text{center}}^0] \)
\end{algorithmic}
\end{algorithm}

In addition to the real-robot experiments, we compare the performance of different functional keypoint transfer strategies from a perception perspective, focusing on the function point transfer. 

\noindent \textbf{Baselines.} We evaluate four function point transfer strategies: 
\begin{itemize}[leftmargin=*]
    \item Demo+VLM+DSC (proposed), which utilizes demonstration functional keypoints as references to prompt the VLM for region proposal, followed by point transfer through a dense semantic correspondence model;
    \item Demo+VLM, which removes the dense semantic correspondence model from the proposed implementation;
    \item Demo+DSC (Robo-ABC), which relies solely on a dense semantic correspondence model for functional keypoint transfer, following the approach in Robo-ABC;
    \item VLM (ReKep), which directly prompts the VLM to propose functional keypoints in a zero-shot manner, as done in ReKep. 
\end{itemize}

\noindent \textbf{Experimental Setup.} For each test tool used in the real-robot experiment, we capture RGB images from 6 different views, covering various positions and orientations within the workspace. Each image has a resolution of 1280*720. A total of 150 images are used for evaluation. 

\noindent \textbf{Evaluation Protocol.} To collect ground truth for function point transfer evaluation,  five volunteers were asked to annotate keypoints on test images using demonstration function points as references. Two evaluation metrics are used: (1) Average Keypoint Distance (AKD), which measures the average pixel distance between ground truth and detected keypoints. (2) Average Precision (AP), which represents the proportion of correctly detected keypoints under various thresholds. AP is evaluated under three thresholds: 15, 30, and 45 pixels. 

\begin{table}[t]
\centering
\renewcommand\arraystretch{1.5}
\setlength\tabcolsep{3pt}
\begin{tabular}{ccccc}
\toprule
\multirow{2}{*}{\textbf{Method}} & \multirow{2}{*}{AKD (pixel) $\downarrow$} & \multirow{2}{*}{AP@15 (\%) $\uparrow$} & \multirow{2}{*}{AP@30 (\%)$\uparrow$} & \multirow{2}{*}{AP@45 (\%)$\uparrow$} \\
                                 &                              &                            &                            &                            \\ \hline
Demo+VLM                         & 26.42                        & 38.89                      & 68.44                      & 83.56                      \\
Demo+DSC                         & 33.54                        & 47.11                      & 68.67                      & 78.67                      \\
VLM               & 56.09                        & 15.56                      & 36.22                      & 52.67                      \\
Demo+VLM+DSC         & \textbf{18.54}                        & \textbf{51.33}                      & \textbf{85.78}                      & \textbf{94.44}                      \\ \bottomrule
\end{tabular}
\label{tab:func_transfer}
  \vspace*{0.1in}
\caption{Quantitative results of function point transfer}
  \vspace*{-0.3in}
\end{table}

\noindent \textbf{Quantitative results.} The quantitative results of function point transfer are presented in Table \ref{tab:func_transfer}. The proposed Demo+VLM+DCS consistently outperforms the ablated strategies in both AKD and AP metrics. Demo+VLM achieves reasonable performance by leveraging the rich commonsense knowledge embedded in VLMs. However, VLMs alone struggle to provide precise point-level correspondences, which limits the effectiveness of Demo+VLM compared to the proposed strategy. Meanwhile, relying solely on the dense semantic correspondence model (i.e., Demo+DSC) often fails when faced with large intra-function variations. The performance gap between Demo+VLM and VLM  highlights the importance of using demonstrations as in-context references for the keypoint proposal.

\subsubsection{Function Frame Construction}

Function frames  $\Pi^{t}_R$ and $\Pi^{t}_H$ are constructed based on the 3D functional keypoints $ K^{t}_H = [p_{\text{func}}^{t}, p_{\text{grasp}}^{t}, p_{\text{center}}^{t}]$ and $ K^{t}_R = [q_{\text{func}}^{t}, q_{\text{grasp}}^{t}, q_{\text{center}}^{t}]$, respectively. $\Pi^{t}_H$ is defined by the following elements:

\begin{enumerate}[leftmargin=*]
    \item \textbf{Function axis} 
    \begin{itemize}
        \item Definition: \[
    \mathbf{v}_H^{t} = \frac{p_{\text{func}}^{t} - p_{\text{center}}^{t}}{\|p_{\text{func}}^{t} - p_{\text{center}}^{t}\|}
    \]
    \item Description:  $\mathbf{v}_H^{t}$  is a normalized vector that defines the function axis. It points from the center point $p_{\text{center}}^{t}$ to the function point $p_{\text{func}}^{t}$ at $t$. This axis represents the principal direction along which the function operates.    
    \end{itemize}
    \item \textbf{Grasp vector}
    \begin{itemize}
        \item Definition: \[  
    \mathbf{u}_H^{t} = \frac{p_{\text{grasp}}^{t} - p_{\text{func}}^{t}}{\|p_{\text{grasp}}^{t} - p_{\text{func}}^{t}\|}
        \]
    \item Description: $\mathbf{u}_H^{t}$  is a normalized vector that points from the function point $p_{\text{func}}^{t}$ to the grasp point $p_{\text{grasp}}^{t}$ at $t$.
    \end{itemize}
    \item \textbf{Unit normal vector}
    \begin{itemize}
        \item Definition:
        \[ 
    \mathbf{n}_H^{t} = \frac{\mathbf{u}_H^{t} \times \mathbf{v}_H^{t}}{\| \mathbf{u}_H^{t} \times \mathbf{v}_H^{t} \|}
        \]
        \item Description: $\mathbf{n}_H^{t}$ is the unit normal vector of the function plane $\mathcal{P}_H^{t}$.
    \end{itemize}
    \item \textbf{Function plane}
    \begin{itemize}
        \item Definition:
        \[   
    \mathcal{P}_H^{t}: (\mathbf{p} - p_{\text{func}}^{t}) \cdot \mathbf{n}_H^{t} = 0
        \]
        \item Description: $\mathcal{P}_H^{t}$ is defined by the function point and its normal vector, describing the tool’s spatial configuration at $t$. 
    \end{itemize}
\end{enumerate}
Similarly, \(\mathbf{v}_R^t\), \(\mathbf{u}_R^t\), \(\mathbf{n}_R^t\), and $\mathcal{P}_R^{t}$ are defined for \( \Pi_R^t \).\\

\subsubsection{Function Frame Alignment}
To enhance the robustness and adaptability of MimicFunc, we introduce a VLM-based state evaluator for semantic refinement in the second stage of function frame alignment. In this stage, MimicFunc first renders the predicted function keyframe interaction by back-projecting the combined point cloud of the test tool and target object onto the camera plane. The rendered scene is then provided as input to the VLM, which assesses whether the predicted state is functionally valid.

\begin{figure}[h]
\centering
\includegraphics[width=1\linewidth]{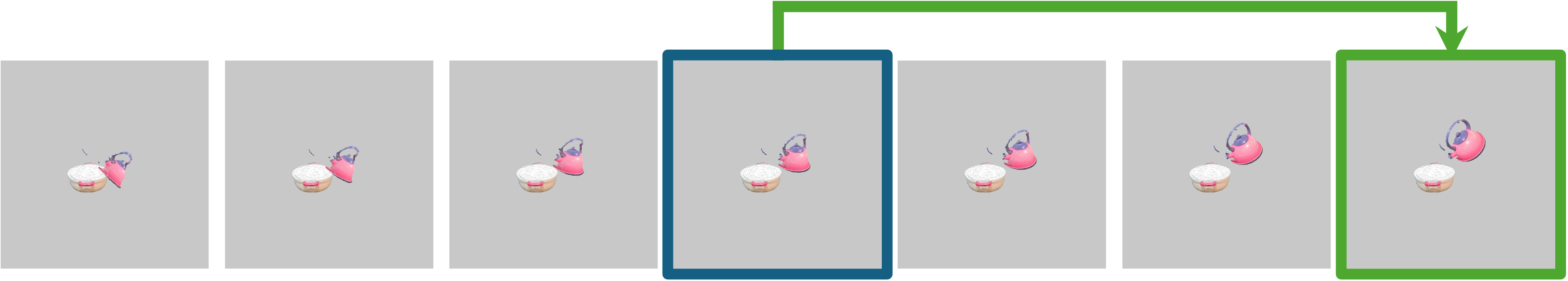}
\caption{Intermediate rendering results of function axis refinement.}
\label{fig:rendering}
\end{figure}

If the state is deemed valid, the alignment is accepted for downstream action generation. Otherwise, the VLM sequentially inspects each primitive to automatically pinpoint those responsible for the failure. Using this feedback, MimicFunc uniformly resamples candidate points or axes around the initial constraint and iteratively repeats the evaluation process until a valid alignment is achieved. Figure \ref{fig:rendering} illustrates an example of intermediate rendering results during function axis refinement, where blue denotes the initial alignment and green indicates the refined result.

\subsection{Function Frame-based Trajectory Generation}\label{sec:appendix_c}

In this section, we provide implementation details for trajectory generation, complementing the constrained optimization problem formulated in the manuscript.  

\noindent \textbf{Trajectory Warping.} Given a demonstration function frame trajectory $\{\boldsymbol{\Pi}_{\text{H}}^t\}_{t=0}^{N-1}$ and its associated function point trajectory $\{p_{\text{func}}^t\}_{t=0}^{N-1}$, trajectory warping adapts these references to a new test scenario by leveraging geometric symmetries and relational transformations. The process consists of three main stages:

\begin{enumerate} [leftmargin=*]
    \item \textbf{Symmetry-Based Repositioning:} \\
    If the target object exhibits geometric symmetries (e.g., rotational symmetry about one of the principal axes), we exploit this property to reposition the demonstration so that the test tool can adopt a more feasible approach direction. Let $R_{\text{sym}} \in \mathrm{SO}(3)$ denote a symmetry rotation. Then, the demonstration function frames and points are transformed as:
    \[
    \boldsymbol{\Pi}_{\text{H}}^{t} = R_{\text{sym}} \cdot \boldsymbol{\Pi}_{\text{H}}^t, \quad 
    p_{\text{func}}^{t} = R_{\text{sym}} \cdot p_{\text{func}}^t
    \]

    \item \textbf{Function Frame Trajectory Pre-processing:} \\
    We pre-process the demonstration’s function frame trajectory by first applying a rotation around one of the principal axes (e.g., $x$-, $y$-, or $z$-axis). The alignment angle $\theta$ is computed based on the angular difference between the initial function points of the demonstration and the test tool:
    \[
    \theta = \angle(p_{\text{func}}^0, q_{\text{func}}^0), \quad 
    R_{\text{align}}(\theta) \in \mathrm{SO}(3)
    \]
    The aligned frame is then obtained by:
    \[
    \boldsymbol{\Pi}_{\text{align}}^t = R_{\text{align}}(\theta) \cdot \boldsymbol{\Pi}_H^t
    \]

    \item \textbf{Function Frame Trajectory Transformation:} \\
    To account for differences in position and scale between the demonstration and the test tool, we apply a translation $\mathbf{t} \in \mathbb{R}^3$ and an optional scaling factor $s \in \mathbb{R}$:
    \[
    \boldsymbol{\Pi}_{\text{warp}}^t = s \cdot \boldsymbol{\Pi}_{\text{align}}^t + \mathbf{t}
    \]
\end{enumerate}

\noindent \textbf{Optimization Constraints and Costs.} 
Beyond the trajectory cost and keyframe constraints detailed in the manuscript, we introduce the following enhancements:

\begin{itemize} [leftmargin=*]
    \item \textbf{Early Trajectory Cost Relaxation.}  
    To encourage smoother transitions and allow flexibility during the approach phase, the trajectory cost is omitted for the initial 30\% of the trajectory. This is particularly beneficial when the initial states of the demonstration and test tools differ significantly, as the primary interaction occurs later in the motion.
    
    \item \textbf{Velocity Constraint.}  
    We constrain both translational and angular velocities of the test tool to ensure smooth motion and physical feasibility throughout the trajectory.

    \item \textbf{Collision Avoidance Constraint.}  
    A minimum Euclidean distance is enforced between the test tool and the 3D bounding box of nearby obstacles, preventing collisions during execution.
\end{itemize}

\noindent We use \texttt{CasADi} for symbolic modeling and automatic differentiation, and solve the resulting nonlinear constrained optimization problem with \texttt{IPOPT}.

\subsection{Data Generation for Visuomotor Policy Training}\label{sec:appendix_d}

\textbf{Data Generation.} To acquire data for visuomotor policy training, we leverage MimicFunc to generate rollout trajectories for novel tools, without requiring labor-intensive teleoperation data collection for novel objects. The process begins with the robot randomly sampling object layouts, including the positions and orientations of the tool, within its workspace. The robot then places the tool at the sampled configuration. MimicFunc then generates a candidate motion to accomplish the task. After executing the rollout, the final scene is captured by the camera and evaluated using a VLM to determine task success. Only successful rollouts are retained to construct a demonstration dataset, which is used to train visuomotor policies capable of generalizing across diverse tools and object arrangements.

\textbf{Policy Training.} We experiment with two state-of-the-art behavioral cloning approaches: Action Chunking Transformer (ACT) and Diffusion Policy (DP), both of which utilize a DINOv2-pretrained ResNet-18 as the visual encoder backbone. ACT has demonstrated effectiveness in learning complex manipulation skills using transformer architectures. In our implementation, we adapt the standard ACT by incorporating RGB-D inputs from two viewpoints: a third-person camera and an in-hand camera. The policy receives the most recent frame from each camera and predicts absolute end-effector poses. This design enables the generation of long-horizon trajectories by predicting action chunks of 100 steps in a single forward pass, making it particularly suitable for tasks that demand precise global positioning. DP takes a generative approach using denoising diffusion probabilistic models. It receives two consecutive RGB-D frames as input and predicts the next 16 delta end-effector poses through an iterative denoising process over 100 inference steps. This approach excels at modeling complex, multimodal action distributions and is well-suited for tasks requiring smooth and continuous motion. The hyperparameters used for both policies are summarized in Table~\ref{tab:policy_hyperparams}.

\begin{table}[h]
\centering
\renewcommand\arraystretch{1.5}
\setlength\tabcolsep{5pt}
\begin{tabular}{lcc}
\toprule
\textbf{Parameter} & \textbf{ACT} & \textbf{DP} \\
\midrule
Visual Encoder & DINOv2-ResNet-18 & DINOv2-ResNet-18 \\
Action Representation & Absolute EE Pose & Delta EE Pose \\
Observation Horizon & 1 & 2 \\
Chunk Size & 100 & 16 \\
Hidden Dimension & 512 & -- \\
Feedforward Dimension & 3200 & -- \\
Encoder Layers & 4 & -- \\
Decoder Layers & 7 & -- \\
Attention Heads & 8 & -- \\
Batch Size & 64 & 64 \\
Epochs & 500 & 500 \\
Learning Rate & 1e-4 & 1e-4 \\
Scheduler & Cosine Annealing & Cosine Annealing \\
KL Weight & 10.0 & -- \\
Diffusion Timesteps & -- & 100 \\
EMA Power & -- & 0.75 \\
\bottomrule
\end{tabular}
\vspace{0.2in}
\caption{Hyperparameters used for ACT and DP training. ``--" indicates the parameter is not applicable.}
\label{tab:policy_hyperparams}
\end{table}

\textbf{Data Quality.} In addition to showcasing the capability of MimicFunc for efficient data generation, we conduct experiments to further evaluate the quality of the generated data. Specifically, we compare the performance of the ACT trained on two different data sources for the \texttt{Pour} task: (1) teleoperation-collected demonstrations, and (2) MimicFunc-generated demonstrations (50 samples). The results show that ACT trained on MimicFunc-generated data achieves a higher success rate (53.85\%) compared to ACT trained on teleoperation data (46.15\%). This performance gain supports our claim that MimicFunc produces more consistent and higher-quality data. In contrast, teleoperation data often suffers from variability, inconsistencies in execution, and imprecision due to human control limitations.

\end{document}